\documentclass[sigconf,screen]{acmart}

\usepackage[font=footnotesize]{caption} 
\usepackage{booktabs}
\usepackage{epsfig}
\usepackage{romannum}
\usepackage{multirow}
\usepackage{diagbox}
\usepackage{pifont}
\usepackage{algorithm}
\usepackage{algpseudocode}

\newcolumntype{P}[1]{>{\centering\arraybackslash}p{#1}}

\AtBeginDocument{%
  }

\setcopyright{acmlicensed}
\copyrightyear{2023}
\acmConference[WWW '23]{Proceedings of the ACM Web Conference 2023}{May 1--5, 2023}{Austin, TX, USA}
\acmBooktitle{Proceedings of the ACM Web Conference 2023 (WWW '23), May 1--5, 2023, Austin, TX, USA}
\acmYear{2023}
\acmPrice{15.00}
\acmDOI{10.1145/3543507.3583874}
\acmISBN{978-1-4503-9416-1/23/04}

\begin{document}

\title{Vertical Federated Knowledge Transfer via Representation Distillation for Healthcare Collaboration Networks}


\author{Chung-ju Huang}
\email{chongruhuang.pku@gmail.com}
\affiliation{%
  \institution{Key Laboratory of High Confidence Software Technologies (Peking University), Ministry of Education}
  \city{Beijing}
  \country{China}
}
\affiliation{%
  \institution{School of Computer Science, Peking University}
  \city{Beijing}
  \country{China}
}

\author{Leye Wang}
\email{leyewang@pku.edu.cn}
\affiliation{%
  \institution{Key Laboratory of High Confidence Software Technologies (Peking University), Ministry of Education}
  \city{Beijing}
  \country{China}
}
\affiliation{%
  \institution{School of Computer Science, Peking University}
  \city{Beijing}
  \country{China}
}

\author{Xiao Han}
\authornote{Corresponding author}
\email{xiaohan@mail.shufe.edu.cn}
\affiliation{%
  \institution{School of Information Management and Engineering, Shanghai University of Finance and Economics}
  \city{Shanghai}
  \country{China}}

\renewcommand{\shortauthors}{Huang et al.}

\begin{abstract}
  Collaboration between healthcare institutions can significantly lessen the imbalance in medical resources across various geographic areas. However, directly sharing diagnostic information between institutions is typically not permitted due to the protection of patients' highly sensitive privacy. As a novel privacy-preserving machine learning paradigm, federated learning (FL) makes it possible to maximize the data utility among multiple medical institutions. These feature-enrichment FL techniques are referred to as vertical FL (VFL). Traditional VFL can only benefit multi-parties' shared samples, which strongly restricts its application scope. In order to improve the information-sharing capability and innovation of various healthcare-related institutions, and then to establish a next-generation open medical collaboration network, we propose a unified framework for \textbf{v}ertical \textbf{fed}erated knowledge \textbf{trans}fer mechanism (VFedTrans) based on a novel cross-hospital representation distillation component. Specifically, our framework includes three steps. First, shared samples' \textit{federated representations} are extracted by collaboratively modeling multi-parties' joint features with current efficient vertical federated representation learning methods. Second, for each hospital, we learn a \textit{local-representation-distilled module}, which can transfer the knowledge from shared samples' federated representations to enrich local samples' representations. Finally, each hospital can leverage local samples' representations enriched by the distillation module to boost arbitrary downstream machine learning tasks. The experiments on real-life medical datasets verify the knowledge transfer effectiveness of our framework.
  \vspace{-0.5em}
\end{abstract}

\ccsdesc[500]{Computing methodologies~Knowledge representation and reasoning}
\ccsdesc[500]{Applied computing~Health care information systems}

\keywords{vertical federated learning, healthcare collaboration network, knowledge distillation, representation learning}


\maketitle

\section{Introduction}

Currently, the disparity in healthcare resources \cite{0004LMAA21} continues to be a significant challenge for both developed and developing countries. For a myriad of reasons, there are huge differences in the healthcare resources accessible to distinct areas, classes, and ethnicities even within the same country \cite{RazaNQRMTAHRV22,barr2019health}. After Covid-19 officially became a pandemic, it overwhelmed healthcare facilities in less developed areas due to the severity of clinical symptoms and the unpredictability of post-recovery sequelae \cite{MaMGJYZRWTW21}. The supply-demand conflict of unbalanced healthcare resources is thus rapidly increasing, which will affect the sustainability of the healthcare system \cite{FattahiKKG23} and the viability of health policy reform \cite{FengWWH21}. In order to promote social equality and social justice, it is crucial from a strategic perspective to address the disparity in healthcare resources \cite{YeCWLXM21,KeyaIPSF21}. In this paper, we will focus on the secure utilization of adequate medical data from developed regions to make up for inadequate and incomplete hospital data from lagging regions.

Unlike other fields, medical data contains many of the most private details of patients' personal lives, psychological conditions, social relationships, and financial situations, making it particularly sensitive to privacy \cite{antunes2022federated,LiuCZY0BJNX022,marazzi2022staying,KushnerS20}. The disclosure of such identifiable privacy about individuals can greatly damage the level of public trust in healthcare institutions. Therefore, no institution will be permitted to provide patient information to another institution directly. With the enactment of the EU General Data Protection Regulation (GDPR)\footnote{\url{https://gdpr-info.eu/}} and US Health Insurance Portability and Accountability Act (HIPAA)\footnote{\url{https://www.hhs.gov/hipaa/for-professionals/privacy/index.html}}, access to and use of private data has been further restricted. How to conduct machine learning and data mining in a privacy-preserving and law-regulated way has attracted much interest in both academia and industry. Federated learning (FL) \cite{mcmahan2017communication} has thus become a promising solution \cite{FFSTXL22,Liu0C21}. In general, FL does not need different parties to exchange their raw data; instead, every party runs local computation and training on their own data and then uploads the intermediate results (e.g., gradients) to a server. By integrating these intermediate results from all the parties, a federated global model can be learned. Especially, such a federated model can achieve similar prediction performance as a centralized model directly trained on all the parties' data \cite{yang2019federated}.

In general, there are two main types of FL algorithms, \textit{horizontal} and \textit{vertical}. The first FL algorithm proposed by Google is horizontal \cite{mcmahan2017communication}; the setting is that different parties (often devices) hold different samples with the same features or data formats. A representative application of horizontal FL is the mobile phone keyboard next-word prediction, where a global next-word prediction model can be learned without collecting users' raw keyboard inputs \cite{yang2018applied}. In contrast, vertical FL's (VFL) setting is that different parties (often organizations) hold different features of the same set of samples. This work focuses on the vertical setting.

The successful adoption of current VFL methods is highly dependent on how many overlapped samples exist between parties. Hence, most VFL collaborations are conducted by involving at least one giant data holder with abundant data. For instance, FDN (federated data network) \cite{LiuFCXY21} includes anonymous data from one of the largest social network service providers in China and thus can cover most user samples from other data holders (e.g., customers of banks). However, this makes giant data holders occupy a dominant position over other small data holders in VFL, which could lead to unfair trades and data monopoly in the digital economy.\footnote{\url{https://www.theguardian.com/technology/2015/apr/19/google-dominates-search-real-problem-monopoly-data}} Collaborative healthcare network \cite{warnat2021swarm} is composed of hospitals with multiple locations and different medical resources. In this scenario, the characteristics and attributes of smaller hospitals' own data are often overlooked when larger hospitals with more data dominate the collaboration. The capacity of the hospitals receiving assistance to use the local data's specificity to give patients more individualized treatment is severely hampered by this. For patients, they may go to different hospitals for the same disease. Differences in the level of care at the hospital will affect the analysis and diagnosis of the disease. High level hospitals tend to detect more hidden symptoms, i.e., have richer sample features. However, these medical records can only be kept in multiple locations. This leads to the fact that in the traditional VFL, patients can only get better joint services if they have been to multiple hospitals. Patients who are limited to a few or even one hospital due to region, race, etc. do not have access to equitable medical resources. Attention and protection for this group are crucial and necessary. To alleviate this pitfall and expand application scenarios, \textit{a VFL-based collaborative framework that can benefit various ordinary hospitals and vulnerable populations is urgently needed}.



As a pioneering attempt in this direction, this paper proposed a novel vertical-federated-knowledge-transfer (VFedTrans) unified framework that can transfer the medical knowledge from (a limited number of) collaborative healthcare networks' shared samples to each hospital's local (non-shared) samples. The key challenge is \textit{how to fill the gap between hospital's local samples (with only this hospital's features) and cross-hospital shared samples (with multiple hospitals' features)}. To address this issue, we propose a novel local-representation-distilled module that can distill the knowledge from shared samples' federated representations to enrich local samples' representations. More specifically, shared samples' federated representations are first learned by some federated latent representation extraction methods (e.g., federated singular vector decomposition \cite{chai2022practical}); then, the small hospital can leverage shared samples' federated representation as the guidance to enrich its local samples' feature representation via a knowledge distilling strategy \cite{Hinton2015DistillingTK}. Especially, our knowledge transfer mechanism has the following characteristics.

\begin{itemize}
  \item \textit{Knowledge transfer to local samples}. Different from most VFL algorithms focusing on shared samples, our mechanism aims to improve the learning performance on different parties' local samples via vertical knowledge transfer. In this way, a set of hospitals with only a limited number of shared samples can still benefit from our VFL process.
  
  \item \textit{Task-independent transfer}. Our knowledge transfer process is task-independent. That is, each hospital can leverage its enriched local samples' representations for an arbitrary (new) medical task. 
  
  \item \textit{Scalable to multiple hospitals}. The complexity of our mechanism is linearly proportional to the number of involved hospitals. More importantly, our mechanism can be learned in an online manner. That is, when a new hospital comes, existing hospitals can efficiently update their local sample representations by just learning with the new hospitals.
\end{itemize}

In summary, this work makes the following contributions:

\begin{enumerate}
  \item To the best of our knowledge, this work is the first one to explore how to enable vertical knowledge transfer from shared samples to each hospital's local samples in a task-independent manner. 

  \item We propose a novel \textit{federated-representation-distilled} framework, \textit{VFedTrans}, to transfer medical knowledge from shared samples to local samples. VFedTrans includes the following steps. First, a federated representation learning method is applied to extract shared samples' representations. Second, each hospital can enrich its local feature representation-distilled module by knowledge distilling on the shared samples' federated representations. The module can then be leveraged to enrich local samples' feature representations.

  \item Experiments on four real-life medical datasets have verified the effectiveness of our mechanism for knowledge transfer and the generalizability of our enriched feature representations of local samples. This demonstrates that VFedTrans enables hospitals with scarce medical resources to provide better medical services through VFL collaboration.
\end{enumerate}

The source code for this work is available at: \url{https://doi.org/10.5281/zenodo.7623519}.

\section{Problem Formulation}
\label{sec:problem}

In this section, we clarify the definitions of key concepts used in this paper. Afterward, we formulate our research problem. Appendix \ref{app:notion} lists the notations used throughout this paper.

\subsection{Concepts}
Our approach enables all hospitals to benefit from the collaboration. Without loss of generality, we classify them into two types of roles: the \textit{task hospital} and the \textit{data hospital}.

\textit{Task Hospital}. A task hospital $t$ has a set of samples with features $X_t$ and a task label $Y_t$ to predict. The sample IDs of the task hospital are denoted as $I_t$. 

\textit{Data Hospital}. A data hospital $d$ has a set of samples with features $X_{d}$. The data hospital's sample IDs are denoted as $I_{d}$. 

\textbf{Remark}. A hospital may play both task and data roles simultaneously in a VFL campaign (i.e., a hospital contributes its features to other hospitals and also benefits from other hospitals' features). Our mechanism can be efficiently applied to this case.

\begin{figure*}
  \centering
  \includegraphics[width=\linewidth]{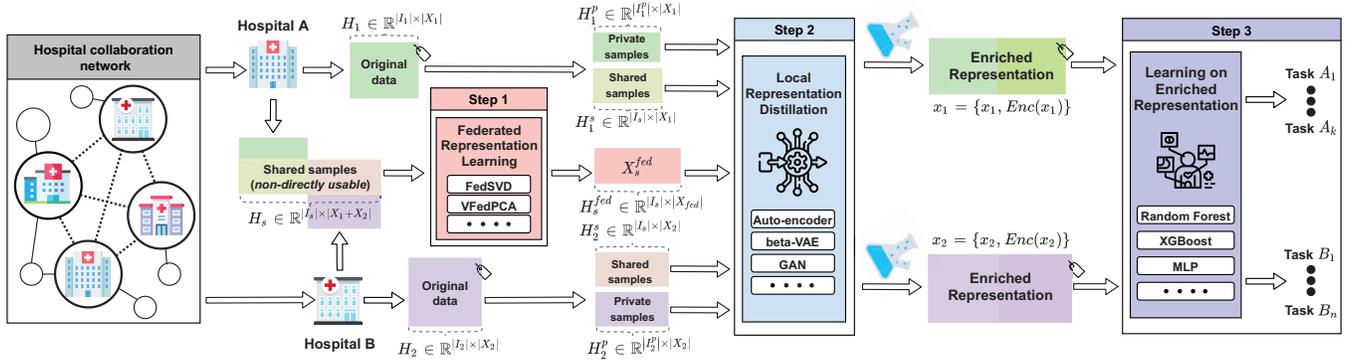}
  \vspace{-2em}
  \caption{Flowchart of VFedTrans for two hospitals. Each hospital can be either the task or data party. For task $A_1,...,A_k$, \textit{Hospital} \textit{A} is the task party and \textit{Hospital} \textit{B} is the data party; for task $B_1,...,B_k$, \textit{Hospital} \textit{B} is the task party and \textit{Hospital} \textit{A} is the data party. In Step 1, we use the VFL techniques for federated representation learning on non-directly usable shared samples to get a federated latent representation $x_s^{fed}$. In Step 2, both hospitals are able to train a local-representation-distilled module for knowledge transfer on their own. For the shared samples, the loss function in knowledge distillation not only contains the reconstruction loss but also adds a new extra distillation loss term that we designed. Then we align the learned representations with the original data to obtain new local enriched representations. In Step 3, both hospitals are able to use the learned enriched representations to complete their respective downstream healthcare-related machine-learning tasks.}
  \label{fig:flowchart}
  \vspace{-4mm}
\end{figure*}

\subsection{Research Problem}
\label{sub:problem}

\textit{Local-Sample Vertical Federated Knowledge Transfer Problem.} Given a task hospital $t$ and $n$ data hospitals $d_i (i=1,2,...,n)$, $t$ has certain shared samples with any data hospital $d_i$ ($I_t \cap I_{d_i} \not = \phi$), 
the objective is to design a federated knowledge transfer algorithm to predict the task label $Y_t$ of $t$'s (non-shared) local samples as accurately as possible. 

\textbf{Remark}. Traditional VFL problems often require that $I_t = I_{d_i}$. However, our vertical federated knowledge transfer setting only needs that $I_t \cap I_{d_i} \not = \phi$. Without loss of generality, we solely validate the impact of knowledge transfer on a task hospital to check that VFedTrans has strong service support for healthcare-related institutions with limited knowledge. Briefly, the objective of our proposed VFedTrans is to improve the task performance of $t$'s local samples ($I_t \setminus I_{d_i}$) by transferring the knowledge from shared samples ($I_t \cap I_{d_i}$). This expands the practical application range of FL by complementing traditional VFL.

\section{Framework Design}

\label{sec:machanism}

\subsection{Overview}

\begin{figure}[t]
  \centering
  \includegraphics[width=\linewidth]{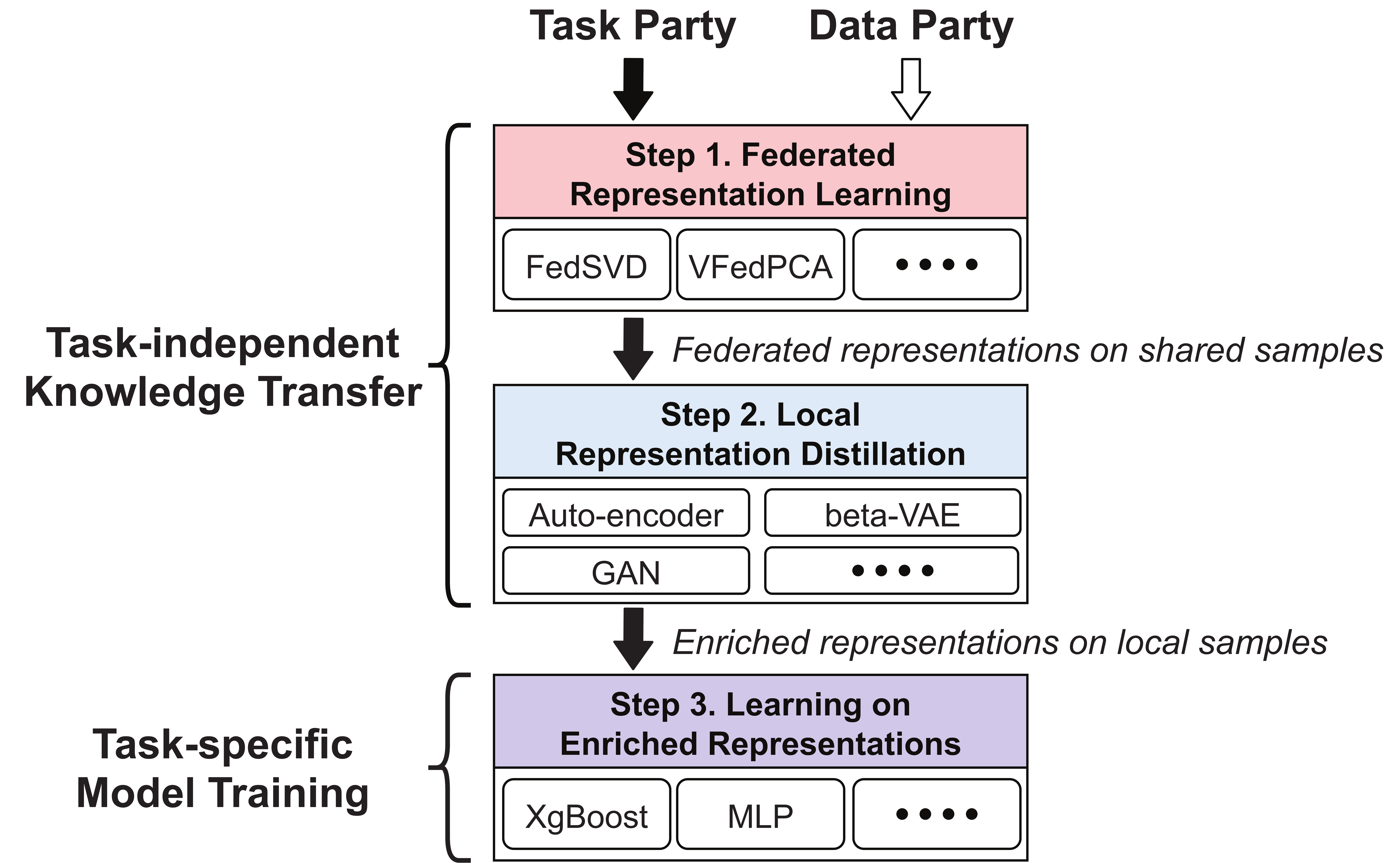}
  \vspace{-2em}
  \caption{Overview of VFedTrans.}
  \label{fig:mechanism_overview}
  \vspace{-1em}
\end{figure}

We demonstrate the overall process of VFedTrans (Fig.~\ref{fig:flowchart}). Note that before our mechanism runs, we suppose that shared samples between the task hospital $t$ and any data hospital $d_i$ are known, which can be learned by PSI (private set intersection) methods \cite{Kamara2014ScalingPS}. Our framework can be simplified into three main steps (Fig.~\ref{fig:mechanism_overview}).

\begin{itemize}
  \item \textbf{Step 1. Federated Representation Learning (FRL)}. First, the task hospital and data hospital collaboratively learn federated latent representations for shared samples using secure VFL techniques. In brief, these federated latent representations would incorporate the hidden knowledge among multiple parties while not leaking these parties' raw features.
	
  \item \textbf{Step 2. Local Representation Distillation (LRD)}. Second, the task hospital trains a \textit{federated-representation-distilled module} that can distill the knowledge from shared samples' federated latent representations to enrich local samples' representations.
	
  \item \textbf{Step 3. Learning on Enriched Representations}. After Step 2, the FRD module is distilled and ready for local feature enrichment. Then, given an arbitrary label $y_t$ to predict, the task hospital can use local samples' enriched representations (i.e., task hospital's local features + enriched representations) to conduct training and inference with state-of-the-art (SOTA) machine learning algorithms.
\end{itemize}

Step 3 generally follows traditional supervised learning methods to train a task-specific medical prediction model, where various machine learning algorithms can be applied, such as random forest and neural networks. Next, we illustrate more details about Step 1 and 2. For brevity, we first assume that only one data hospital $d$ exists. At the end of Sec.~\ref{subsec:lrd}, we will discuss how to deal with multiple data hospitals $\{d_1, d_2,...,d_n\}$.

\subsection{Federated Representation Learning}
\label{subsec:frl}

The purpose of Step 1 is to extract latent representations of shared samples by considering both task and data hospitals' features. In general, we can adopt various vertical federated representation learning methods for this step. 
In this work, we adopt a matrix decomposition-based federated representation method, as literature has shown that matrix decomposition is effective for extracting meaningful latent representations for machine learning tasks \cite{kosinski2013private}.
Here, we introduce how to leverage two state-of-the-art federated matrix decomposition methods, i.e., FedSVD \cite{chai2022practical} and VFedPCA \cite{cheung2022vertical}, 
to learn shared samples' federated representations by considering both task and data hospitals' features.

\subsubsection{FedSVD}
In FedSVD \cite{chai2022practical}, all hospitals use two random orthogonal matrices to transform the local samples into local masked samples. This maintains the invariance of the decomposition results despite the masking transformation of the local samples. The masked samples are then uploaded to a third-party server, which applies the SVD algorithm to the samples from all hospitals. Finally, the task hospital can reconstruct the federated latent representation based on the decomposition results.

Suppose the task hospital holds the shared samples' feature matrix $S_t \in \mathbb R^{|I_s|\times|X_t|}$, and the data hospital $d$ holds the shared samples' feature matrix $S_d \in \mathbb R^{|I_s|\times|X_d|}$ ($I_s = I_t \cap I_d $ is the shared sample ID set). Denote $S = [S_t|S_d]$ (combination of both task and data hospitals' feature matrices), we want to leverage $S = U\Sigma V^T$ (SVD) to learn the latent representations $U$,
Inspired by FedSVD \cite{chai2022practical}, we use a randomized masking method to learn $U$ as,
\begin{enumerate}
  \item A trusted key generator generates two randomized orthogonal matrices $A\in \mathbb R^{|I_s|\times|I_s|}$ and $B \in \mathbb R^{|X_{td}|\times|X_{td}|}$ ($|X_{td}| = |X_t| + |X_d|$). $B$ is further partitioned to two parts $B_t \in \mathbb R^{|X_t|\times|X_{td}|}$ and $B_d \in \mathbb R^{|X_d|\times|X_{td}|}$, i.e., $B^T = [B_t^T | B_d^T]$.
	
  \item $A$ and $B_t$ are sent to the task hospital; $A$ and $B_d$ are sent to the data hospital. Each hospital does a local computation by masking its own feature matrices with the received matrices:
  \begin{equation}
    \hat S_k = A S_k B_k, \forall k \in \{t,d\}
  \end{equation}

  \item Both task and data hospitals send $\hat S_t$ and $\hat S_d$ to a third-party server\footnote{The third-party server needs to be semi-honest. Note that in FL, such a security configuration (i.e., the information aggregation server is semi-honest) is widely accepted \cite{yang2019federated}.} and the third-party server runs SVD on the combined data matrix $\hat S = \hat U \Sigma \hat V^T$, where $\hat S = [\hat S_t|\hat S_d]$.
  $\hat U$ is then sent to the task hospital.
	
  \item The task hospital can recover the federated latent representation of shared samples, denoted as ${\bf x}_s^{fed}$, by
  \begin{equation}
    {\bf x}_s^{fed} = U = A^T \hat U
  \end{equation}
\end{enumerate}

Compared to the original FedSVD which aims to recover both $U$ and $V$ \cite{chai2022practical}, we only need to recover $U$. Hence,  in VFedTrans, only $\hat U$ is transmitted to the task hospital to reduce the communication cost. The correctness of the above process depends on the fact that $S$ and $\hat S$ (multiplying $S$ by two orthogonal matrices) must hold the same singular value $\Sigma$ \cite{chai2022practical}. 

\subsubsection{VFedPCA}
To enhance the generality of VFedTrans, we also use vertical federated principal component analysis (VFedPCA) \cite{cheung2022vertical} to extract latent representations. Under VFedPCA's setting, each hospital makes its own federated eigenvector $u$ converge to global eigenvector $u_G$ without needing to know the mutual data of all hospitals. Each hospital is able to train the local eigenvector using local power iteration \cite{saad2011numerical}. Then the eigenvectors from each hospital are merged into the federated eigenvector $u$. Finally, task hospital can use $u$ to reconstruct the original data to obtain the federated latent representation.

Suppose the task hospital holds the shared samples' feature matrix $S_t \in \mathbb R^{|I_s|\times|X_t|}$, and the data hospital holds the shared samples' feature matrix $S_d \in \mathbb R^{|I_s|\times|X_d|}$. Denote $S = [S_t|S_d], S \in \mathbb R^{|I_s|\times |X_t+X_d|}$.

\begin{enumerate}
    \item For each hospital $i \in \{t,d\}$, we calculate the largest eigenvalue $A_i=\frac{1}{|X_i|}S_i^T S_i$ and a non-zero vector $a_i$ corresponding to the eigenvector $\alpha_i(A_ia_i=\alpha_i a_i)$. The number of local iterations is $L$, each hospital will compute locally until convergence as follows:
    \begin{equation}
        a_i^{l}=\frac{A_i a_i^{l-1}}{||A_i a_i^{l-1}||}, \quad \alpha_i^l=\frac{A_i{(a_i^l)}^T a_i^l}{{(a_i^l)}^T a_i^l}
    \end{equation}
    where $l = 1, 2, \cdots, L$.
    \item Then each hospital upload the eigenvector $a_i^L$ and the eigenvalue $\alpha_i^L$ to third-party server. The server aggregates the results and generates the federated eigenvalue:
    \begin{equation}
        u = w_t a_t^L + w_d a_d^L,\quad w_i = \frac{\alpha_i^L}{\sum_{i \in \{t, d\}}\alpha_i^L}
    \end{equation}
    \item Task hospital $t$ can use the federated eigenvalue $u$ to reach the federated latent representation:
    \begin{equation}
        {\bf x}_s^{fed} = S_t\frac{MM^T}{||MM^T||},M=S_t^T u
    \end{equation}
\end{enumerate}

\subsection{Local Representation Distillation}
\label{subsec:lrd}

After obtaining ${\bf x}^{fed}_s$ for shared samples, Step 2 aims to enrich the task hospital's local sample representations. We thus design a novel local feature extracting strategy, which is combined with knowledge distilling from shared samples' ${\bf x}^{fed}_s$. Specifically, for a certain unsupervised local representation learner, we enhance it by adding a new loss function, i.e., making the shared samples $I_s$'s learned representations be close to ${\bf x}^{fed}_s$, thus enabling the knowledge distillation effect.

\begin{figure}[t]
  \centering
  \includegraphics[width=.75\linewidth]{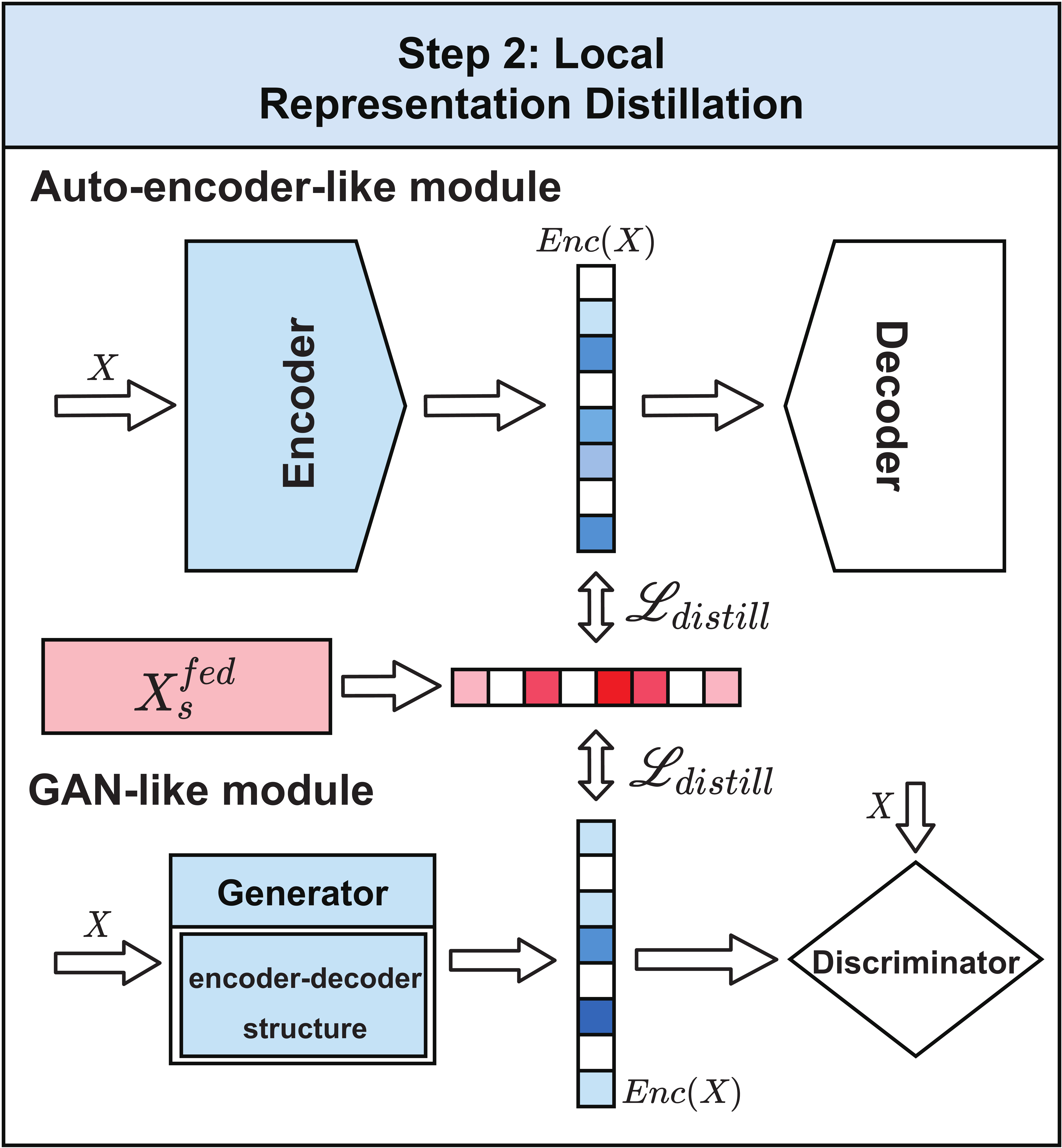}
  \vspace{-1em}
  \caption{The structure of distillation module in Sec. \ref{subsec:lrd}. In VFedTrans, the generator of the GAN-like module uses the encoder-decoder structure \cite{Tran0017,PangWCCW21}.}
  \label{fig:LRD}
  \vspace{-2em}
\end{figure}

In our mechanism implementation, we consider several representative unsupervised representation extraction methods, i.e., auto-encoder (AE) \cite{Hinton2006ReducingTD}, beta-VAE \cite{higgins2016beta}, and GAN \cite{goodfellow2020generative}. Especially, if the input features are from a shared sample, we add a knowledge distillation loss function by comparing the encoder's output to the shared sample's federated representation (learned from Step 1),
\begin{equation}
  \mathcal L_{distill}({\bf x}_s^t) = |Enc({\bf x}_s^t) - {\bf x}_s^{fed}|
\end{equation}
where ${\bf x}_s^t$ is the shared samples' local features in the task hospital. Fig. \ref{fig:LRD} shows the structure of distillation module for various representation extraction methods.

Hence, the complete loss function of the distilled module is,
\begin{equation}
\label{equa:loss}
  loss = 
  \begin{cases}
    \mathcal L_{recons}({\bf x}_i^t) + \theta \mathcal L_{distill}({\bf x}_i^t)  & i \in I_s \\
    \mathcal L_{recons}({\bf x}_i^t) & i \in I_t \setminus I_s
  \end{cases}
\end{equation} 
That is, for the task hospital's (non-shared) local samples, the loss function is the same as the original distillation module. For the shared samples, a new knowledge distillation loss is added to the original reconstruction loss; $\theta$ is the weight parameter to balance two loss function parts.

After training the federated-representation-distilled module until convergence, the distillation module's output can be a feature enrichment function for the task hospital's local samples. That is, for $i \in I_t \setminus I_s$, $Enc({\bf x}_i^t)$ can be used to enrich the original local feature ${\bf x}_i^t$. In other words, the enriched local samples' representations ${\bf x}_i^* = \langle {\bf x}_i^t, Enc({\bf x}_i^t) \rangle$ are given to Step 3 for training a machine learning model for a medical task.

\textbf{Extension to multiple data hospitals}. When there are $n$ data hospitals, the task hospital can repeat the aforementioned Step 1 and 2 with each data hospital. Specifically, for each data hospital $d_i$, the task hospital can learn a local feature enrichment function $Enc_i$. Then, by aggregating $n$ local feature enrichment functions learned from $n$ data hospitals, the local samples' final enriched representations become,
\begin{equation}
  {\bf x}_i^* = \langle {\bf x}_i^t, Enc_1({\bf x}_i^t), Enc_2({\bf x}_i^t), ..., Enc_n({\bf x}_i^t) \rangle
\end{equation}

Appendix~\ref{app:algorithm} summarizes the pseudo-code of VFedTrans.

\subsection{Security and Privacy}

Security and privacy are key factors to consider in FL mechanism design.
While VFedTrans is a knowledge transfer framework that incorporates existing VFL algorithms, the security and privacy protection levels are mainly dependent on the included VFL algorithm. In particular, the FRL module (Sec.~\ref{subsec:frl}) is the key part to determine the overall security and privacy levels of VFedTrans, as cross-party communications and computations are only conducted in this step.
Currently, we implement the FRL module with SOTA VFL representation learning methods including, FedSVD \cite{chai2022practical} and VFedPCA \cite{cheung2022vertical}.FedSVD uses two random orthogonal matrices to mask the original data. The third-party server can only use the masked data of each party to obtain SVD result. The third-party server of VFedPCA only needs to use the eigenvectors and eigenvalues of each party's data for weighted summation. All of these methods protect privacy by preventing direct use of data by non-holders. Due to the page limitation, readers may refer to the original papers \cite{chai2022practical,cheung2022vertical} for specific security and privacy analysis.

\subsection{Updating}
\label{sub:updating}

In general, VFedTrans is efficient to update without the need to completely re-running three steps for all the hospitals.

  \textit{\textbf{Local Incremental Learning} - New task hospital samples}. Note that the representation distillation is conducted locally at the task hospital. Then, if the task hospital $t$ has a number of new local samples, $t$ can locally re-conduct the representation distillation to learn an updated local feature enrichment function. The task hospital $t$ does not need to communicate with any data hospitals for this updating, which is very efficient and convenient.\footnote{In practice, for new samples, the existing local feature enrichment function (without updating) can still be used. Our evaluation would test this setting (Sec. \ref{sec:inductive}).}
  
  \textit{\textbf{Task Independence} - New tasks}. Similar to new samples, if the task hospital $t$ has a new task label to predict, $t$ also does not need to communicate with other hospitals. $t$ only needs to repeat Step 3 with the new task label.
  
  \textit{\textbf{Knowledge Extensibility} - New data hospitals}. 
  The task hospital can learn a new local feature enrichment function $Enc'$ from the new data hospital (repeat Steps 1 and 2 with the new data hospital), and then enrich the local feature representation as ${\bf x}_i^* = \langle {\bf x}_i^*, Enc'({\bf x}_i^t) \rangle $.

\section{Evaluation}

In this section, we empirically verify the effectiveness of our mechanism with four real-life medical datasets. Our experiments were performed on the workstation using NVIDIA RTX 3090, Intel(R) Xeon(R) Gold 6330 CPU @ 2.00GHz, 160GB RAM, PyTorch 1.10.0, Python 3.8 and CUDA 11.3.

\subsection{Datasets}
\label{subsec:datasets}
We evaluate our mechanism on the four medical-related datasets: MIMIC-\Romannum{3} \cite{johnson2016mimic}, RNA-Seq \cite{weinstein2013cancer}, HAPT \cite{reyes2016transition}, and Breast \cite{breast}. Appendix \ref{app:dataset} details these four datasets and shows the default data split to different hospitals in the experiments. We suppose that there exists one task hospital and one data hospital by default. Due to the page limitation, for most experiments, we show the results on MIMIC-\Romannum{3} and HAPT datasets.

\subsection{Baselines}

To verify the effectiveness of our mechanism, we compare it with four baselines:

\begin{itemize}
  \item \textit{LOCAL}: This baseline leverages only the task hospital's local features for training the task-specific model.
  \item \textit{FTL \cite{Liu2020ASF}}: FTL is an end-to-end FL method for transferring knowledge to local samples. Specifically, based on shared samples, FTL maps different parties' raw features to a common feature space to achieve knowledge transfer.
  \item \textit{IAVFL \cite{RenYC22}}: IAVFL first learns a federated model on shared samples and then learns a local model (for local samples) by considering both ground-truth labels and soft labels produced by the federated model.
  \item \textit{FedSimLoc}: FedSim \cite{fedsimloc} is originally designed for fuzzy linking of samples between VFL parties when samples' exact identifies are unavailable. We modify it to our scenario, denoted as \textit{FedSimLoc}, by linking a task party's local (non-shared) sample with the top-$K$ similar shared samples. Then, this local sample can be predicted considering its similar shared samples' federated features from other data parties.
  
\end{itemize}

While FTL, IAVFL, and FedSimLoc can be used to assist local samples' learning in VFL, they do not hold some key characteristics of VFedTrans, such as \textit{task-independence} (Sec.~\ref{sub:updating}). 
Moreover, these baselines are all designed purely using deep learning models (i.e., neural networks), whereas VFedTrans can do prediction tasks using any machine learning model (e.g., random forest and XGBoost).
Note that in many medical tasks, traditional machine learning models still perform very efficiently and effectively \cite{morrill2019signature} (we also run a set of experiments on our datasets to verify this in Sec.~\ref{sub:robustness}); VFedTrans is thus more suitable for such medical tasks.



\subsection{Training Configurations}
\label{subsec:training}
In our experiments, we use the random forest (RF) as the default machine learning algorithm, FedSVD as the default FRL method, and AE as the default LRD method. Details of the remaining training configurations are in Appendix \ref{app:training}.

\subsection{Main Results}
\label{sub:main}

\begin{figure*}[t]
	\centering
	\begin{minipage}{0.32\linewidth}
		\centering
		\includegraphics[width=1\linewidth]{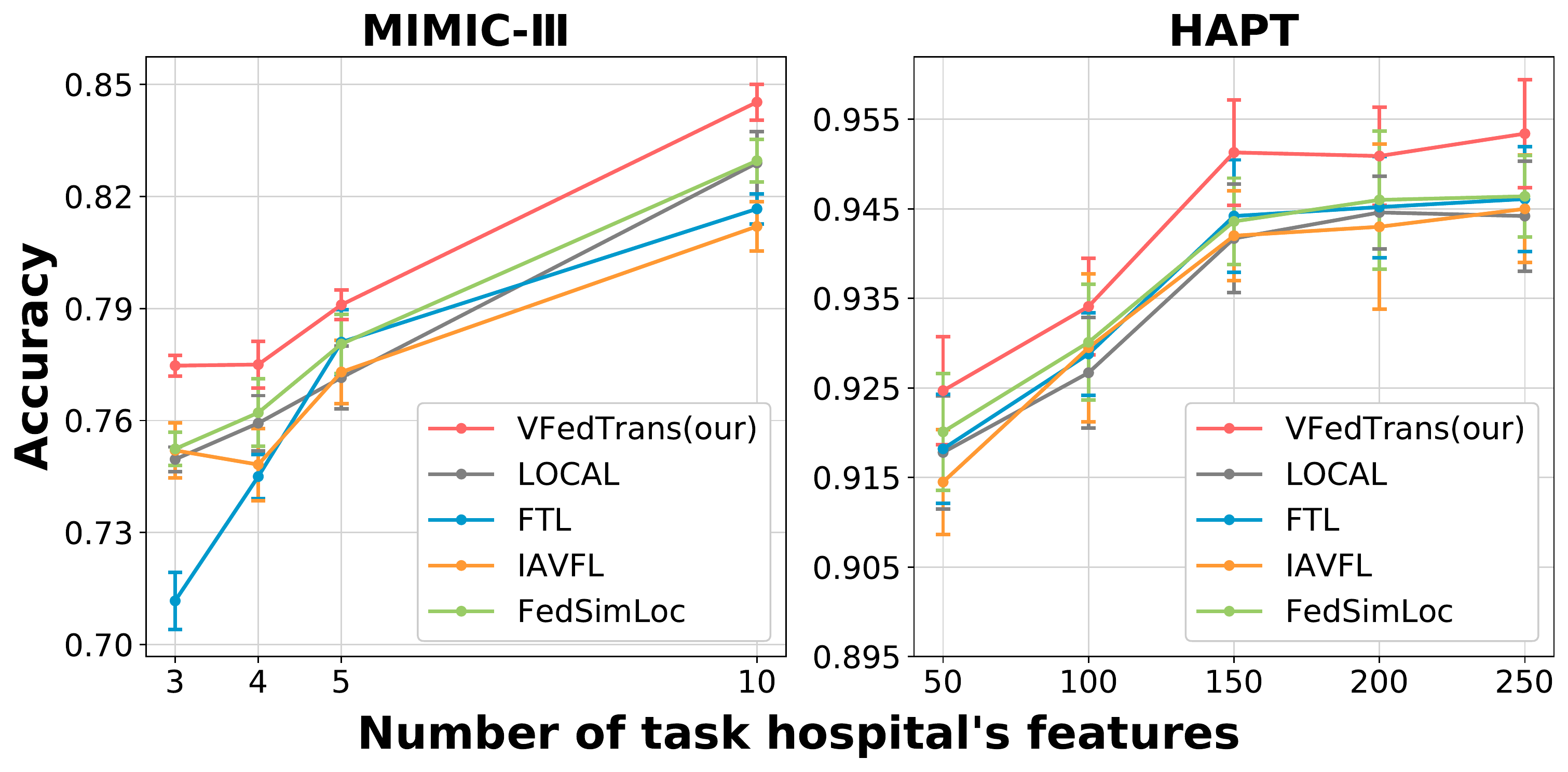}
		\vspace{-2.5em}
		\caption{Prediction accuracy by varying the task hospital's feature number.}
		\label{fig:vary_task_feature}
	\end{minipage}
	\begin{minipage}{0.32\linewidth}
		\centering
		\includegraphics[width=1\linewidth]{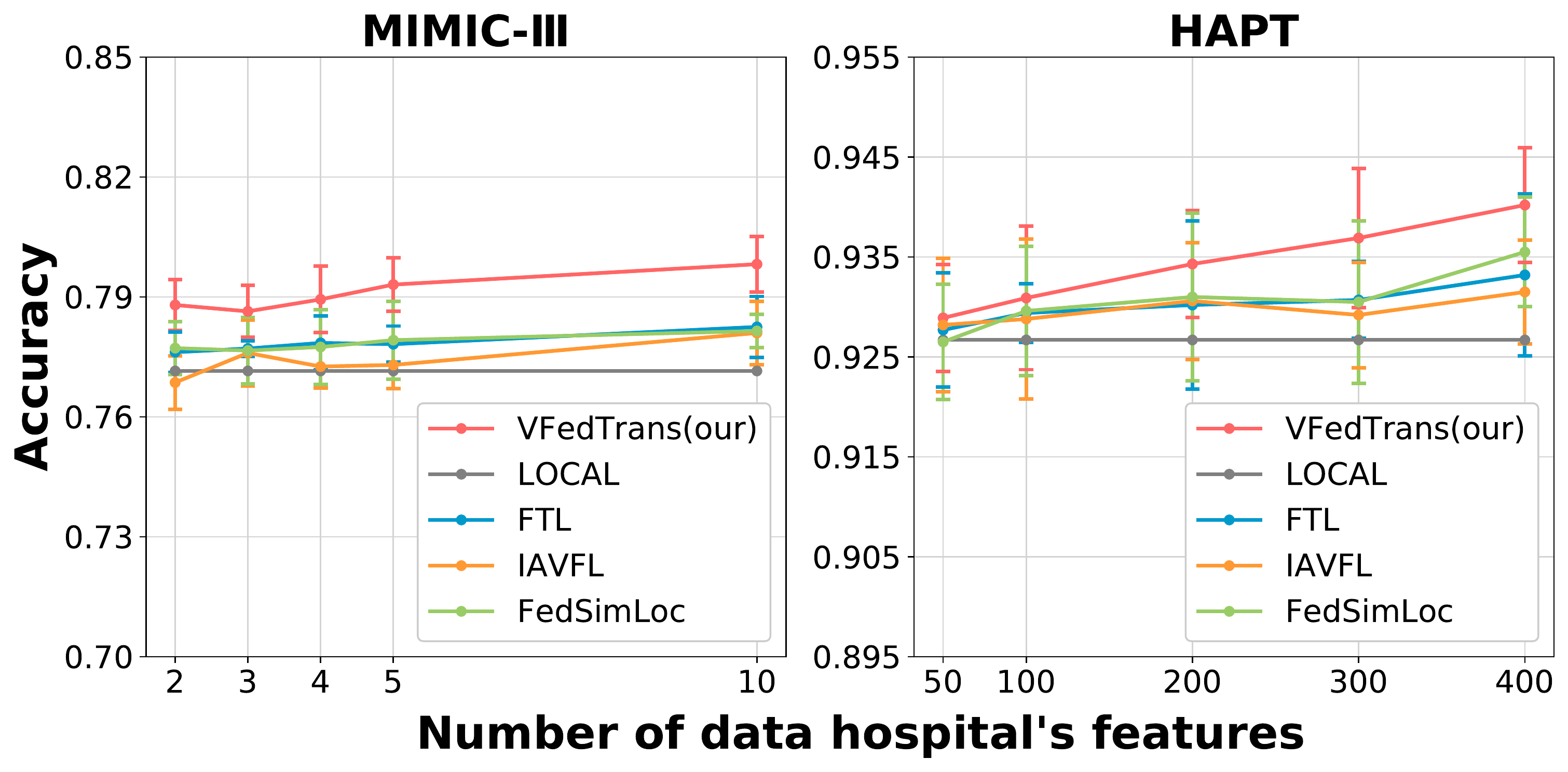}
		\vspace{-2.5em}
		\caption{Prediction accuracy by varying the data hospital's feature number.}
		\label{fig:vary_data_feature}
	\end{minipage}
	\begin{minipage}{0.32\linewidth}
		\centering
		\includegraphics[width=1\linewidth]{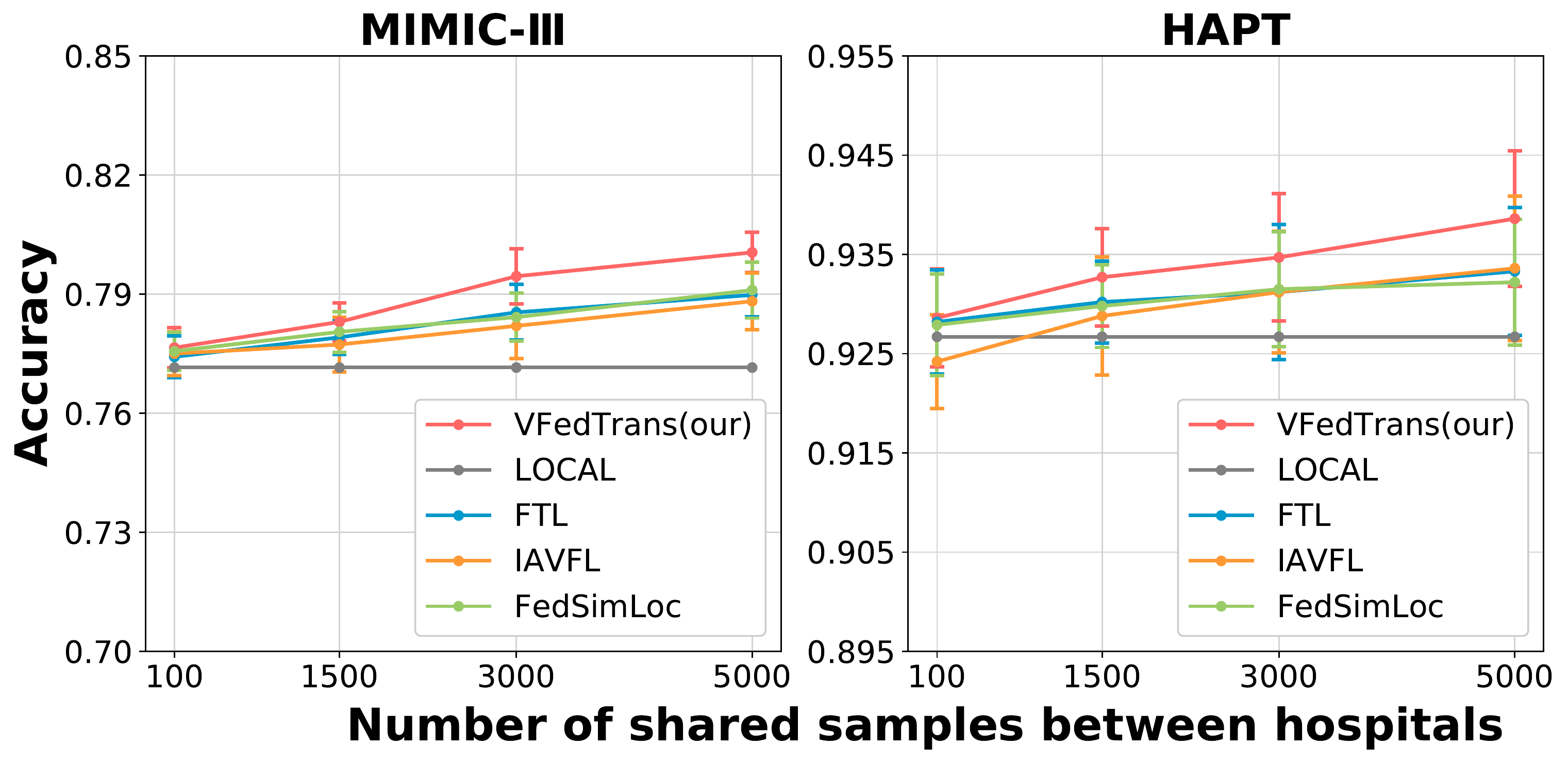}
		\vspace{-2.5em}
		\caption{Prediction accuracy by varying the number of shared samples.}
		\label{fig:vary_shared_sample}
	\end{minipage}\hfill
	\begin{minipage}{0.32\linewidth}
		\centering
		\includegraphics[width=1\linewidth]{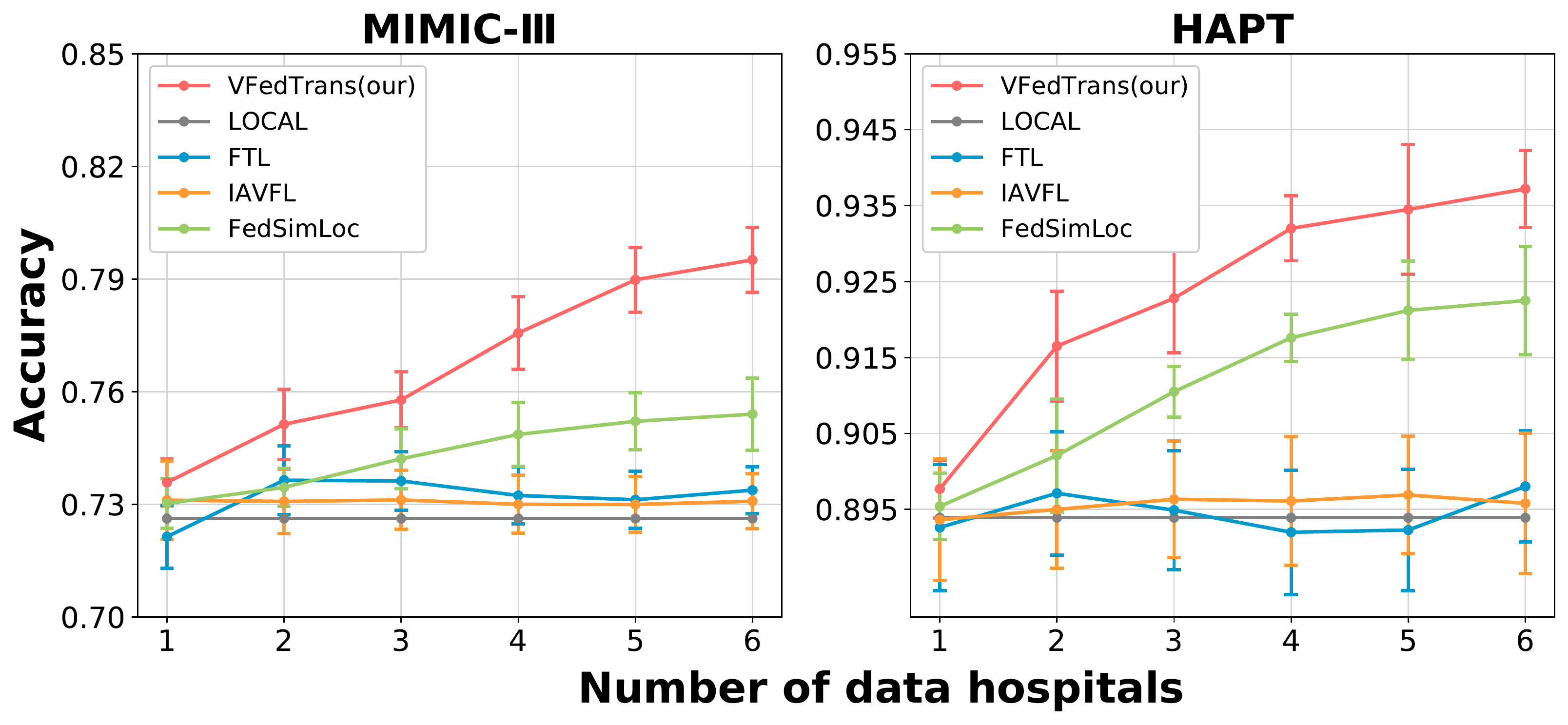}
		\vspace{-2.5em}
		\caption{Prediction accuracy by varying the number of data hospitals.}
		\label{fig:vary_party}
	\end{minipage}
        \begin{minipage}{0.32\linewidth}
            \centering
            \vspace{-0.75em}
            \includegraphics[width=1\linewidth]{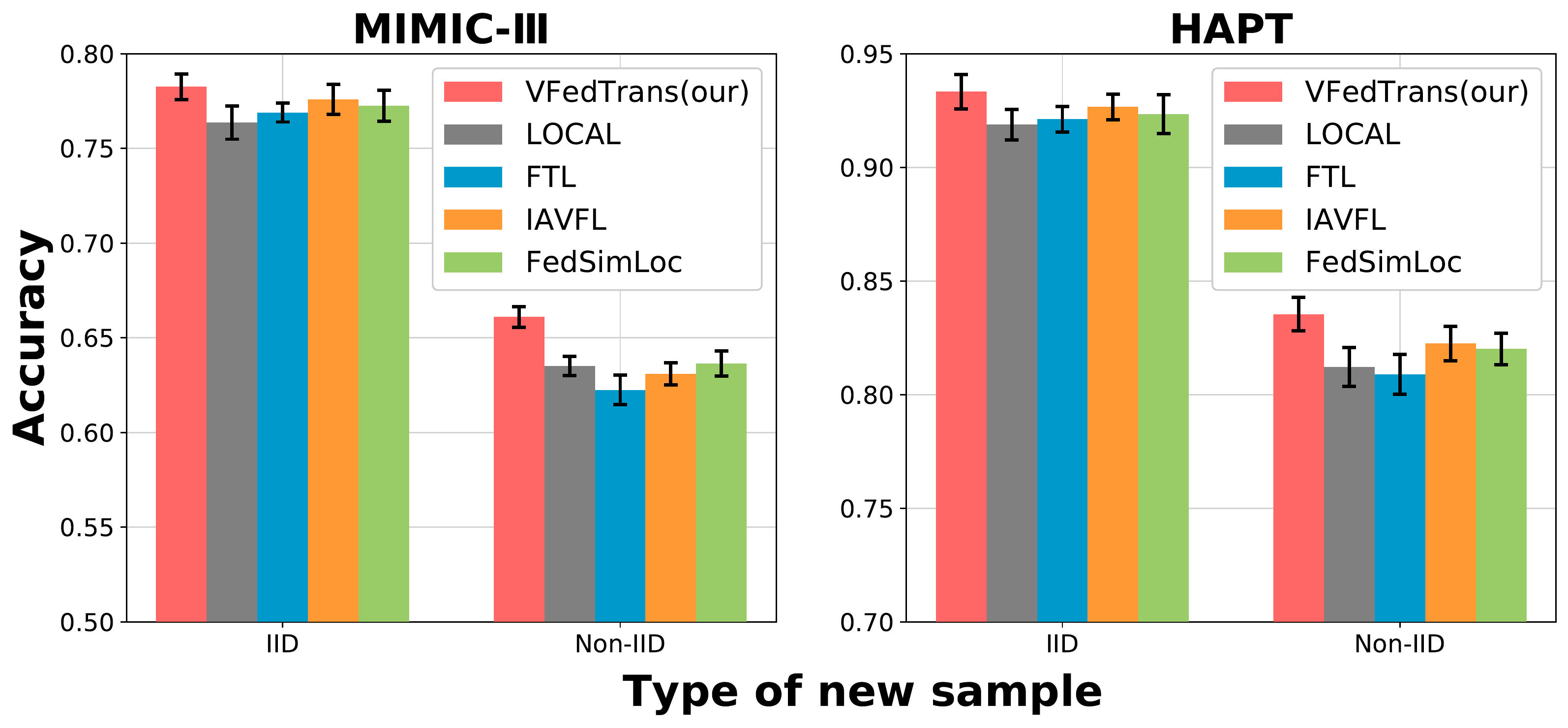}
            \vspace{-2.5em}
            \caption{Prediction accuracy of new samples.}
            \label{fig:inductive_new_sample}
        \end{minipage}
        \begin{minipage}{0.32\linewidth}
            \centering
            \vspace{0.5em}
            \includegraphics[width=1\linewidth]{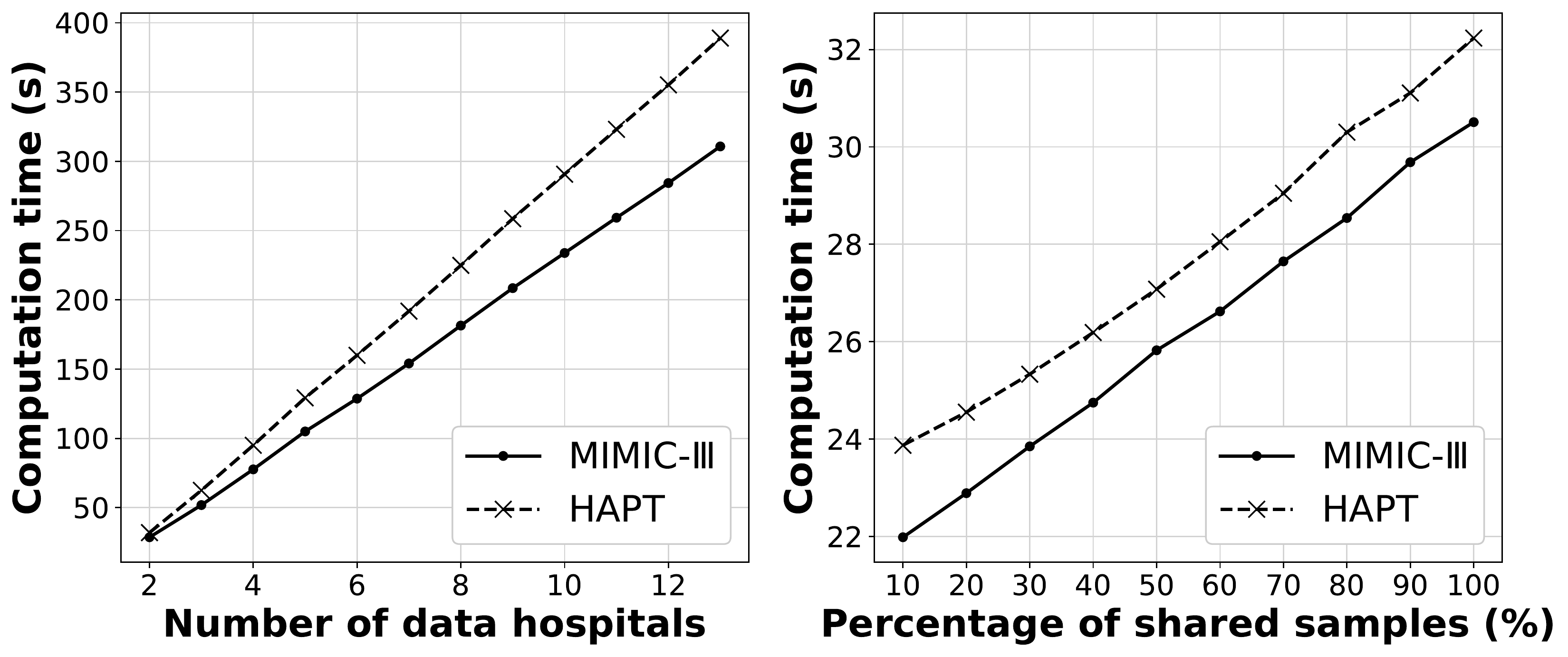}
            \vspace{-2.25em}
            \caption{Computation time by varying the number of data parties and the number of shared samples.}
            \label{fig:computation}
        \end{minipage}\hfill
\end{figure*}

We first report the results when there is one data hospital. Fig. \ref{fig:vary_task_feature} depicts the prediction performance on MIMIC-\Romannum{3} and HAPT by varying the number of features in the task hospital. Results show that our framework can consistently outperform LOCAL and FL baselines. The former means that VFedTrans can effectively improve the diagnosis accuracy when the task hospital's features are insufficient; the latter shows that our knowledge distillation mechanism can achieve better performance compared to other FL knowledge transfer mechanisms.

Fig. \ref{fig:vary_data_feature} shows the prediction performance on MIMIC-\Romannum{3} and HAPT by varying the number of features in the data hospital. The accuracy of VFedTrans gradually goes up as the number of features in the data hospital increases. More interestingly, the accuracy increasing speed of VFedTrans is more significant than baselines. 
This indicates that VFedTrans can transfer knowledge from rich features of the data hospital much more efficiently than baselines. 


Fig.~\ref{fig:vary_shared_sample} shows how our mechanism performs by changing the number of shared samples between the task hospital and the data hospital. 
We observe that the performance gets better as there are more shared samples.
This also validates the effectiveness of VFedTrans: with more knowledge sources (i.e., shared samples), our transfer can always be better.

\subsection{Few-shot Results}
\label{subsec:fewshot}
Real-world data holders mostly keep unlabeled data and have access to only few samples of labeled data. 
We thus consider a test situation where the task hospital does not have enough labeled samples available. We reduce the samples used for training the downstream model in Step 3 to 10\% of the original size and keep the test part the same. Fig.~\ref{fig:vary_party} explores our mechanism's performance under this few-shot setting. 
When the number of data hospitals is 1 (the same setting as in Sec. \ref{sub:main}), our mechanism can still outperform baselines; meanwhile, all the methods have some degree of performance loss due to the limited number of training samples. 

Additionally, building a VFL mechanism by involving multiple data parties is realistically advantageous, particularly for few-shot cases.
To validate the effectiveness of VFedTrans under this scenario, we increase the number of data hospitals involved in collaboration. 
Note that the baselines FTL and IAVFL do not consider multiple data parties in their original design. 
For these two methods, we run the two-party collaboration between the task hospital and every data hospital, and finally output the ensemble prediction with averaging.
With the increase in data hospitals, the performance of VFedTrans grows obviously and outperforms baselines consistently. 
This means that, with VFedTrans, the task hospital can obtain effective information from multiple data hospitals to compensate for the insufficiency of local data volume and features. 


\subsection{Robustness Check}
\label{sub:robustness}

\begin{table}[t]
	\centering
	\footnotesize
	\begin{tabular}{lcccc}
		\toprule
		\textbf{Method} & \textbf{HAPT} & \textbf{RNA-Seq} & \textbf{MIMIC-\Romannum{3}} & \textbf{Breast} \\ \midrule
		\textit{VFedTrans (ADA)} & $.9002 \pm .0048$ & $.9556 \pm .0034$ & $.6848 \pm .0049$ & $.9233 \pm .0054$ \\
		\textit{LOCAL (ADA)} & $.8825 \pm .0068$ & $.9333 \pm .0052$ & $.6769 \pm .0048$ & $.9067 \pm .0048$ \\ \midrule
		\textit{VFedTrans (NN)} & $\textbf{.9530} \pm .0089$ & $.9600 \pm .0190$ & $.7687 \pm .0119$ & $.8436 \pm .0035$ \\
		\textit{LOCAL (NN)} & $.9502 \pm .0125$ & $.9583 \pm .0172$ & $.7643 \pm .0080$ & $.8250 \pm .0038$ \\ \midrule
		\textit{VFedTrans (KNN)} & $.9203 \pm .0079$ & $.9578 \pm .0158$ & $.6839 \pm .0141$ & $.8583 \pm .0049$ \\
		\textit{LOCAL (KNN)} & $.9122 \pm .0070$ & $.9512 \pm .0102$ & $.6761 \pm .0167$ & $.8300 \pm .0058$ \\ \midrule
		\textit{VFedTrans (XGB)} & $.9519 \pm .0066$ & $.9625 \pm .0140$ & $\textbf{.8042} \pm .0107$ & $.9233 \pm .0043$ \\
		\textit{LOCAL (XGB)} & $.9495 \pm .0057$ & $.9548 \pm .0089$ & $.7940 \pm .0072$ & $.9116 \pm .0087$ \\ \midrule
		\textit{VFedTrans (RF)} & $.9341 \pm .0054$ & $\textbf{.9635} \pm .0036$ & $.7910 \pm .0040$ & $\textbf{.9253} \pm .0068$ \\
		\textit{LOCAL (RF)} & $.9267 \pm .0062$ & $.9524 \pm .0042$ & $.7715\pm .0084$ & $.9100 \pm .0042$ \\ \midrule
		\textit{FTL} & $.9288 \pm .0046$ & $.9413 \pm .0075$ & $.7810 \pm .0087$ & $.8628 \pm .0133$ \\
		\textit{IAVFL} & $.9295 \pm .0083$ & $.9512 \pm .0065$ & $.7735 \pm .0086$ & $.9085 \pm .0066$ \\
		\textit{FedSimLoc} & $.9301 \pm .0065$ & $.9468 \pm .0082$ & $.7805 \pm .0079$ & $.8786 \pm .0084$ \\
		\bottomrule
	\end{tabular}
	\caption{Prediction accuracy on four datasets under different downstream classification models (ADA: AdaBoost, NN: neural networks, KNN: K nearest neighbours, XGB: XGBoost, RF: random forest).}
	\vspace{-7mm}
	\label{tab:results_other}
\end{table}
\begin{table}[t]
	\centering
	\footnotesize
	\begin{tabular}{lcccc}
		\toprule
		\multirow{2}{*}{\textbf{Method}} & \multirow{2}{*}{\textbf{FRL}} & \multirow{2}{*}{\textbf{LRD}} & \multicolumn{2}{c}{\textbf{Accuracy}} \\ 
		& & & MIMIC-\Romannum{3} & HAPT \\ \midrule
		\multirow{6}{*}{\textit{VFedTrans}} & FedSVD & AE & $\textbf{.7910} \pm .0040$ & $.9341 \pm .0054$ \\
		& FedSVD & beta-VAE & $.7895 \pm .0078$ & $\textbf{.9345} \pm .0039$ \\
		& FedSVD & GAN & $.7889 \pm .0058$ & $.9325 \pm .0058$ \\
		& VFedPCA & AE & $.7886 \pm .0045$ & $.9330 \pm .0068$ \\
		& VFedPCA & beta-VAE & $.7875 \pm .0061$ & $.9321 \pm .0034$ \\
		& VFedPCA & GAN & $.7868 \pm .0088$ & $.9332 \pm .0077$ \\ \midrule
		\textit{LOCAL} & - & - & $.7715 \pm .0084$ & $.9267 \pm .0062$ \\
		\textit{FTL} & - & - & $.7810\pm .0087$ & $.9288 \pm .0046$ \\
		\textit{IAVFL} & - & - & $.7735\pm .0086$ & $.9295 \pm .0083$ \\
		\textit{FedSimLoc}  & - & - & $.7805\pm .0079$ & $.9301 \pm .0065$ \\
		\bottomrule
	\end{tabular}
	\caption{Prediction accuracy by changing FRL and LRD modules.}
	\vspace{-9.5mm}
	\label{tab:frl_lrd}

\end{table}

Previous experiments were conducted with the default configurations of VFedTrans and two datasets, MIMIC-\Romannum{3} and HAPT.
Here, we check the robustness of VFedTrans by modifying its configurations (classification, FRL, and LRD modules) on more datasets. 

Table \ref{tab:results_other} shows the results when VFedTrans uses different machine learning models for training the task classifier on four datasets. 
Our VFedTrans consistently outperforms LOCAL and other baselines, verifying the generalized effectiveness of our knowledge transfer method in various datasets and classifier models. 
We also find that no single classification model dominates across all the datasets.
Then, the flexibility of VFedTrans to incorporate any classification model turns out to be a significant benefit in reality, as we can customize the classifier according to the target dataset.

Besides, we change the methods in FRL and LRD of VFedTrans. Table~\ref{tab:frl_lrd} illustrates the prediction accuracy when we modify the modules in VFedTrans. We can see that the resultant accuracy is robust to such modifications.


\subsection{Inductive Learning Results}
\label{sec:inductive}


Moreover, we check how VFedTrans can facilitate inductive learning for new samples of the task hospital (i.e., not used in training the federated-representation-distilled module in Sec.~\ref{subsec:lrd}). In reality, new samples may have a different feature or label distribution from old samples since many factors may change with time, leading to a non-IID case. We thus run inductive learning on both IID and non-IID cases.\footnote{For the non-IID case, the label distribution of new samples is different from training samples. Details are in Appendix \ref{app:inductive}.}

Fig.~\ref{fig:inductive_new_sample} demonstrates that VFedTrans can achieve better performance than other baselines for both IID and non-IID new samples. This reveals the good generalizability of our mechanism's enriched representations. Specifically, while the prediction accuracy decreases for all methods when the experiment setting changes from IID to non-IID, the loss of accuracy is much smaller for our method. For hospitals, the non-IID sample is a completely different case from the original local sample. Models using only local knowledge cannot fit well with a large number of new non-IID samples. Our framework enables these local hospitals to benefit from collaborative federated medical knowledge learning and to maintain a more solid and trustworthy medical diagnosis in the face of unknown cases and more complicated clinical situations.

\begin{figure*}[t]
    \centering
    \includegraphics[width=1\linewidth]{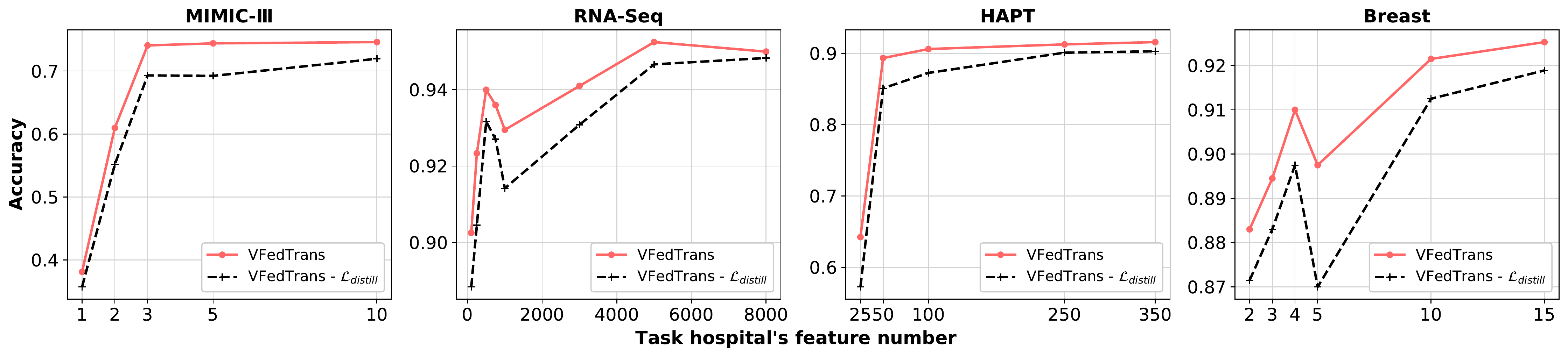}
    \vspace{-2.5em}
    \caption{Prediction accuracy by varying the task hospital's feature number with or without the knowledge distillation loss $\mathcal L_{distill}$.}
    \label{fig:viability}
    \vspace{-1.5em}
\end{figure*}

\subsection{Analysis on $\mathcal L_{distill}$}

To further verify the effectiveness of VFedTrans on knowledge transfer, we compare the changes in prediction accuracy before and after using our proposed knowledge distillation loss function $\mathcal L_{distill} = |Enc({\bf x}_s^t) - {\bf x}_s^{fed}|$ (Sec.~\ref{subsec:lrd}). As shown in Fig. \ref{fig:viability}, the prediction accuracy is significantly improved with the use of the novel loss component for knowledge transfer. Similarly, when the features of the task hospital change, our mechanism always performs better than the mechanism without knowledge transfer. It is worth noting that our proposed loss function can bring higher accuracy improvement when the task hospital has fewer features. This indicates that VFedTrans enables hospitals with insufficient features to significantly benefit from cross-institution collaboration. 

Overall, the results show that the proposed loss is effective for carrying out knowledge transfer. That is, adding $|Enc({\bf x}_s^t) - {\bf x}_s^{fed}|$ into the loss function of representation learning can achieve sensible knowledge transfer. ${\bf x}_s^{fed}$ can be regarded as the teacher and $Enc({\bf x}_s^t)$ as the student. Students use both their own data and the teacher's shared ${\bf x}_s^{fed}$ to conduct knowledge distillation. The generated representations thus benefit from the teacher's knowledge.


\subsection{Computation Time}
\label{subsec:comput}
The computation efficiency of VFedTrans generally depends on the FRL module, as only this step requires collaboration between data and task hospitals.
It is worth noting that our implemented FRL algorithm, such as FedSVD, is highly efficient and can be applied to a billion-scale feature matrix \cite{chai2022practical}.
This fundamentally supports the high computation efficiency of VFedTrans. 

Besides, we vary the problem scale to check how computation time changes.
Fig. \ref{fig:computation} records the computation time of VFedTrans by varying the number of data hospitals and the number of shared samples, respectively. In general, the computation time of our mechanism is linearly proportional to the number of data hospitals and the number of shared samples. This linear relationship further indicates the good scalability of our mechanism. 


\section{Related Work}

In this section, we introduce the related work from two perspectives, VFL methods and FL in healthcare.


\subsection{Vertical Federated Learning}

VFL focuses on cross-organizational collaborative learning. The common setting \cite{yang2019federated} is that different organizations hold different features of the same set of samples. 
Researchers have proposed diverse mechanisms, such as tree-based models \cite{Cheng2019SecureBoostAL,wu2020,abs-2110-10927} and neural networks \cite{hu2019fdml,KangLL22,FuMJXC22} for VFL collaborations. Compared to these VFL algorithms, the key difference of our mechanism is the application scope. Existing VFL algorithms focus on improving the prediction performance on shared samples. In contrast, our mechanism aims to improve the prediction performance of each party's local (non-shared) samples by transferring the knowledge from shared samples. This study validates the effectiveness of our framework for collaborative healthcare learning.
We believe that our proposed VFedTrans can be a good complement to existing VFL algorithms, thus boosting the practicability of FL in reality.

A prior study close to our research is the FTL (federated transfer learning) framework \cite{Liu2020ASF}. FTL first trains a specific neural network model according to the task, and maps the heterogeneous feature space of both parties to a common latent subspace by aligned samples\cite{LiuKang2021}. The task hospital then trains the local network model in this subspace. However, our knowledge transfer process is \textit{task-independent}, which means that the distilled representation of the task hospital's samples (i.e., learned from the distilled encoder) can benefit an arbitrary machine learning task and flexibly select any classification model for local tasks; in comparison, FTL is a neural network-based end-to-end training framework that lacks modules for directly training intermediate layers. This makes it challenging to use non-neural network classifiers in FTL.
Another recent work on local samples' learning for VFL is proposed by \citet{RenYC22}, which transfers the knowledge from the shared samples' federated model to the local model by distilling the soft labels generated by the federated model. Like FTL, it also works in a task-dependent manner based on neural networks, which is different from our VFedTrans.



\subsection{Federated Learning in Healthcare}

FL is a distributed AI paradigm that has been recognized as a promising solution in the field of intelligent healthcare \cite{dayan2021federated,XuGSWBW21,WuCZZ22}.
FedHealth \cite{fedhealth} is designed for wearable health monitoring with smartphone collaboration. Actually, the three steps of this method — local training, model sharing, and server aggregation — are carried out under HFL. Then transfer learning is used in the phase of model personalization. FGTF \cite{FGTF} investigates enhancing a tensor factorization-based collaborative model to handle sensitive health data. FGTF is more concerned with ensuring model convergence and quality reliability while reducing uplink communication costs. Flop \cite{flop} is an application of HFL in the field of medical image classification. In Flop, the client only needs to share a partial model with the server for federated averaging; the remaining few layers of neural network can remain private. These existing federated healthcare frameworks rarely consider how to address the imbalance, insufficiency, and heterogeneity of health data among healthcare institutions from a VFL perspective, which however is the objective of our research. 

\section{Conclusion}

In this work, we propose a vertical-federated-knowledge-transfer unified framework (VFedTrans) to transfer the knowledge from cross-institutional shared samples to each hospital's local samples. VFedTrans can significantly improve the application scenarios of VFL in healthcare collaborative learning, as it is complementary to the traditional VFL solutions that work only for shared samples. Extensive experiments on medical datasets verify the effectiveness of VFedTrans. Future work may focus on incorporating various SOTA FL techniques \cite{Li2021ModelContrastiveFL, blindfl} into VFedTrans to enrich the framework and considering a new blockchain-based peer-to-peer collaborative learning paradigm \cite{warnat2021swarm,YuanHTLY00021,han2022demys} to remove the reliance on the third-party server for higher privacy protection.


\begin{acks}
This work is partially supported by the NSFC Grants no. 72071125, 61972008, and 72031001.
\end{acks}

\bibliographystyle{ACM-Reference-Format}
\bibliography{reference}

\clearpage
\appendix

\section{Appendix}

\subsection{Notation}
\label{app:notion}

The used notation can be found in Table \ref{tab:notion} (as referred to Sec. \ref{sec:problem}.)

\renewcommand{\arraystretch}{1.3}
\begin{table}[h]
  \centering
  \footnotesize
  \begin{tabular}{@{}cc@{}}
    \toprule
    \textbf{Notation} & \textbf{Description} \\ \midrule
    \textbf{$t,d$} & Task hospital and Data hospital. \\
    \textbf{$d_i$} & The i-th data hospital. \\
    \textbf{$X_t,X_d$} & Features of task hospital and data hospital. \\
    \textbf{$I_t,I_d$} & Samples of task hospital and data hospital. \\
    \textbf{$Y_t,Y_d$} & Labels of task hospital and data hospital. \\
    \textbf{$H_i$} & The original local data of hospital $i$. \\
    \textbf{$H_i^p,H_i^s$} & Hospital i's private samples and shared samples. \\
    \textbf{$H_s$} & Non-directly usable shared samples between two hospitals. \\
    \textbf{$x_s^{fed}$} & Federated latent representations (Sec. \ref{subsec:frl}). \\
    \textbf{$Enc(x)$} & Encoder's output in LRD module (Sec. \ref{subsec:lrd}). \\
    \textbf{$x^*$} & Enriched loacal samples' representations. \\
    \textbf{$\mathcal{L}_{recons}$} & Reconstruction loss (Equa. \ref{equa:loss}). \\
    \textbf{$\mathcal{L}_{distill}$} & Novel distillation loss (Equa. \ref{equa:loss}). \\
    \textbf{$\theta$} & Weight parameter (Equa. \ref{equa:loss}). \\ \bottomrule
  \end{tabular}
  \caption{List of used notions.}
  \vspace{-10mm}
  \label{tab:notion}
\end{table}

\subsection{Algorithm}
\label{app:algorithm}

The procedure steps of VFedTrans can be found in Algorithm \ref{alg} (as referred to Sec. \ref{sec:machanism}).

\begin{algorithm}[H]
    \caption{VFedTrans algorithm}
    \label{alg}
    \begin{algorithmic}
        \Require Hospital series: $\{H_1, H_2, \cdots, H_n\}$
        \State Choose two hospitals $H_i$ and $H_j$
        \State Get the private samples $H_i^p$ and $H_j^p$
        \State Get the shared samples $H_i^s$ and $H_j^s$
        \State Get the non-directly usable samples $H_s$
        \State $X_s^{fed} \gets FRL(H_s)$  \Comment{Details are in Sec. \ref{subsec:frl}}
        \For{$H_k \in \{H_i, H_j\}$}
            \State $Enc(X_k) \gets LRD(H_k^p, H_k^s, X_s^{fed})$ \Comment{Details are in Sec. \ref{subsec:lrd}}
            \State $X_k^{*} \gets <X_k,Enc(X_k)>$
        \EndFor
        \For{$H_k \in \{H_i, H_j\}$}
            \State $MedicalTask(X_k^{*})$
        \EndFor
    \end{algorithmic}
\end{algorithm}

\subsection{Dataset}
\label{app:dataset}

This is a supplement to Sec. \ref{subsec:datasets}.

\subsubsection{Details of the medical datasets}

\begin{itemize}
  \item \textit{Medical Information Mart for Intensive Care (MIMIC-\Romannum{3})} dataset provides de-identified health-related data for 58976 patients from 2001 to 2012. The dimension is $\mathbb{R}^{58976 \times 15}$. Length of stays (LOS) is the target of prediction and varies between 1 and 4.
  \item \textit{Human Activities and Postural Transitions (HAPT)} is an activity recognition dataset based on smartphone sensor readings.
  \item \textit{Gene Expression Cancer RNA-Seq (RNA-Seq)} dataset includes gene expressions in patients with different types of tumor. The dimension is $\mathbb{R}^{801 \times 20531}$ and there are 5 tumor types to predict. 
  The dataset dimension is $\mathbb{R}^{10929 \times 561}$. The task label is the activity type (12 types).
  \item \textit{Breast} is calculated from the digital image of the fine needle aspirate of the breast lumps. It has 569 samples and 31 features. The diagnosis result is a binary classification (M = malignant, B = benign).
\end{itemize}

\subsubsection{Data split}

The data held by hospitals under different task settings is shown in Table \ref{tab:data_split}. We shuffle the data before dividing to prevent interference from the label distribution. Assuming that the task hospital $t$ has the fewer resources and its data is insufficient, i.e., $t$ holds a smaller number of samples of features than a data hospital $d$ with more abundant medical resources. This setting is effective in experiments to verify that the knowledge transfer in VFedTrans can better help the weaker party. All experiments except Sec. \ref{subsec:fewshot} and the first experiment in Sec. \ref{subsec:comput} are single-party tasks (one data hospital). In the multi-party task, the number of samples and features of each data hospital are randomly generated within a given interval. The sample size of the shared $H_{s_i}$ is also generated in this manner while the number of features is $X_t+X_{d_i}$. In addition, we set the dimension of the federated latent representation \textbf{$x_s^{fed}$} in Sec. \ref{subsec:frl} to be the same as $H_t$, i.e., both $\mathbb{R}^{I_t \times X_t}$.

\renewcommand{\arraystretch}{1.3}
\begin{table}[h]
  \centering
  \footnotesize
  \begin{tabular}{@{}lcccccc@{}}
    \toprule
    \multirow{2}{*}{\textbf{Task}} &  \multicolumn{2}{c}{\multirow{2}{*}{\textbf{Hospital's data}}} & \multicolumn{4}{c}{\textbf{Dimension}} \\ \cline{4-7}
    & & & \textit{MIMIC-\Romannum{3}} & \textit{HAPT} & \textit{RNA-Seq} & \textit{Breast} \\ \midrule
    \multirow{2}{*}{Common} & \multirow{2}{*}{$H_t$} & $I_t$ & $5000$ & $4000$ & $600$ & $300$ \\
    & & $X_t$ & $5$ & $100$ & $6000$ & $15$ \\ \midrule
    \multirow{4}{*}{Single party} & \multirow{2}{*}{$H_d$} & $I_d$ &$20000$ & $8000$ & $600$ & $400$ \\
    & & $X_d$ &$10$ & $250$ & $8000$ & $15$ \\
    & \multirow{2}{*}{$H_s$} & $I_s$ & $4000$ & $3000$ & $500$ & $200$ \\
    & & $X_s$ & $15$ & $350$ & $16000$ & $30$ \\ \midrule
    \multirow{4}{*}{Multi-party} & \multirow{2}{*}{$H_{d_i}$} & $I_{d_i}$ & $8000\sim25000$ & $5000\sim10000$ & - & - \\
    & & $X_{d_i}$ & $5\sim10$ & $200\sim400$ & - & - \\
    & \multirow{2}{*}{$H_{s_i}$} & $I_{s_i}$ & $2000\sim4000$ & $2000\sim4000$ & - & - \\
    & & $X_{s_i}$ & $X_t+X_{d_i}$ & $X_t+X_{d_i}$ & - & - \\
    \bottomrule
  \end{tabular}
  \caption{Default samples and features held by each hospital.}
  \vspace{-6mm}
  \label{tab:data_split}
\end{table}

\subsection{Training configuration}
\label{app:training}

This is a supplement to Sec. \ref{subsec:training}.

\subsubsection{FRL techniques}
In FRL, we use two VFL techniques, FedSVD and VFedPCA, with the former being the default. We list the key parameters of two methods in Table \ref{tab:frl_setting}.

\begin{table}[h]
  \centering
  \footnotesize
  \begin{tabular}{ccc@{}P{5cm}@{}}
    \toprule
    \textbf{VFL} &\textbf{Parameter} & \textbf{Default} & \textbf{Description} \\ \midrule
    \multirow{2}{*}{FedSVD} &\textit{num\_party} & $2$ & The number of participants. \\
    &\textit{block\_size} & $100$ & Build fix-size block in orthogonal matrix generation. \\ \midrule
    \multirow{4}{*}{VFedPCA} &\textit{party\_num} & $2$ & The number of participants. \\
    &\textit{iter\_num} & $100$ & The number of local power iteration. \\
    &\textit{period\_num} & $10$ & The number of communication period. \\
    &\textit{warm\_start} & True & Use the previous global aggregation vector. \\
    \bottomrule
  \end{tabular}
  \caption{Default key parameters in FRL.}
  \vspace{-10mm}
  \label{tab:frl_setting}
\end{table}

\subsubsection{LRD modules}
We choose Adam optimizer for training distillation module, with learning rate=0.001, batch size=100, epoch=500. AE is the default LRD method. Simultaneously, we also carry out experiments under different LRD modules, such as beta-VAE and GAN, to verify the framework's robustness. The key parameters of the three distillation modules are shown in Table \ref{tab:lrd_setting}.

\begin{table}[h]
  \centering
  \footnotesize
  \begin{tabular}{cccP{4.2cm}@{}}
    \toprule
    LRD &\textbf{Parameter} & \textbf{Default} & \textbf{Description} \\ \midrule
    \multirow{3}{*}{AE} & \textit{depth} & 6 & The depth of encoder and decoder. \\
    & \textit{activation} & Sigmoid & The activation function of hidden layers. \\
    & \textit{$\theta$} & $0.001$ & The weight parameter of $\mathcal{L}_{ditill}$. \\ \midrule
    \multirow{3}{*}{beta-VAE} & \textit{$\beta$} & $4$ & Balance $\mathcal{L}_{recons}$ and $\mathcal{L}_{KL}$. \\
    & \textit{kld\_weight} & $0.00025$ & The weight of $\mathcal{L}_{KL}$. \\
    & \textit{$\theta$} & $0.00001$ & The weight parameter of $\mathcal{L}_{ditill}$. \\\midrule
    \multirow{5}{*}{GAN} & \textit{d\_depth} & $4$ & The depth of discriminator. \\
    & \textit{g\_depth} & $4$ & The depth of generator. \\
    & \textit{activation} & LeakyReLU & The depth of generator and discriminator. \\
    & \textit{negative\_slope} & $0.2$ & The angle of the negative slope. \\
    & \textit{$\theta$} & $0.00001$ & The weight parameter of $\mathcal{L}_{ditill}$. \\
    \bottomrule
  \end{tabular}
  \caption{Default key parameters in LRD.}
  \vspace{-2em}
  \label{tab:lrd_setting}
\end{table}

\subsubsection{Task-specific medical models}
For all the datasets, when training the task-specific medical model, we choose 80\% of the data as the training set and 20\% as the test set. In order to prevent the interference of random seeds, we carry out experiments under 10 different random seeds and compute the average results. Note that we can leverage various machine learning algorithms to train the task-specific model. In our experiments, we use the random forest (RF) as the default machine learning algorithm. We also test the other popular algorithms, including AdaBoost, KNN, XGBoost \cite{Chen2016XGBoostAS}, and neural network (NN) for robustness checks. The parameters of the downstream model are summarized in Table \ref{tab:ml_setting}.

\begin{table}[h]
  \centering
  \footnotesize
  \begin{tabular}{ccc@{}P{4.07cm}@{}}
    \toprule
    \textbf{Model} & \textbf{Parameter} & \textbf{Default} & \textbf{Description} \\ \midrule
    \multirow{2}{*}{RF} & \textit{n\_estimators} & $200$ & The number of the trees. \\
    & \textit{max\_depth} & $10$ & The maximum depth of the tree. \\ \midrule
    \multirow{3}{*}{AdaBoost} & \textit{max\_depth} & $3$ & DecisionTreeClassifier's maximum depth. \\
    & \textit{n\_estimators} & $100$ & The maximum number of estimators. \\
    & \textit{learning\_rate} & $0.5$ & Each classifier's weight at each iteration. \\ \midrule
    \multirow{4}{*}{NN} & \textit{hidden\_layer\_sizes} & $(100, 100, 50)$ & The number of units in hidden layers. \\
    & \textit{$\alpha$} & $0.01$ & Weight of the L2 regularization term. \\
    & \textit{max\_iter} & $400$ & Maximum of iterations. \\
    & \textit{activation} & relu & Activation function for the hidden layer. \\ \midrule
    KNN & \textit{n\_neighbors} & $8$ & Number of neighbors.  \\ \midrule
    \multirow{2}{*}{XGBoost} & \textit{max\_depth} & $7$ & The maximum depth of a tree. \\
    & \textit{learning\_rate} & $0.01$ & Weight at each iteration. \\
    \bottomrule
  \end{tabular}
  \caption{Default key parameters in downstream medical models.}
  \label{tab:ml_setting}
\end{table}

\subsection{Inductive learning: non-IID setting}
\label{app:inductive}
This is a supplement to Sec. \ref{sec:inductive}. We purposely choose half of the labels and their corresponding samples. Then we randomly select 40\% of this portion as new non-IID samples. The remaining 60\% of the samples and the other half of label's samples are used for the generation of hospital data. 

\end{document}